%% file: __main.tex
\pdfoutput=1

\documentclass[11pt]{article}

\usepackage[defaultlines=2,all]{nowidow}

\input{packages}

\newtcolorbox{boxA}{
    enhanced, breakable,
    boxrule = 0pt,
    colback = sub,
    borderline west = {2pt}{0pt}{main}, 
    borderline east = {2pt}{0pt}{main}, 
    width=1.0\linewidth, halign=left, colframe=black, colback=white, boxsep=0.01mm, arc=1.5mm, left=2mm, right=2mm, boxrule=0.5pt
}

\theoremstyle{plain}

\theoremstyle{definition}

\theoremstyle{remark}

\makeatletter
\newcommand\palatino{%
  \def\cmtt@Scale{0.85}
  \fontfamily{cmtt}\selectfont
}
\makeatother
\definecolor{codegreen}{rgb}{0,0.6,0}
\definecolor{codegray}{rgb}{0.5,0.5,0.5}
\definecolor{codepurple}{rgb}{0.58,0,0.82}
\definecolor{backcolour}{rgb}{0.95,0.95,0.95}

\lstdefinestyle{mystyle}{
    backgroundcolor=\color{backcolour},   
    basicstyle=\palatino\footnotesize,
    breakatwhitespace=true,         
    breaklines=true,                 
    captionpos=b,                    
    keepspaces=false,                 
    numbers=none,                    
    numbersep=5pt,                  
    showspaces=false,                
    showstringspaces=false,
    showtabs=false,                  
    tabsize=2
}

\DeclareMathSizes{10}{9}{6}{5}

\title{TinyAgent: Function Calling at the Edge}

\author{Lutfi Eren Erdogan$^{*1}$ \enskip\enskip Nicholas Lee$^{*1}$ \enskip\enskip Siddharth Jha$^{*1}$  \enskip\enskip \bf Sehoon Kim$^{1}$ \\ \bf Ryan Tabrizi$^{1}$ \enskip\enskip Suhong Moon$^{1}$ \enskip\enskip Coleman Hooper$^{1}$\\ \bf Gopala Anumanchipalli$^{1}$ \enskip\enskip Kurt Keutzer$^{1}$ \enskip\enskip Amir Gholami$^{1,2}$\\
$^{1}$UC Berkeley\qquad$^{2}$ICSI \\
{{\small{\{lerdogan, nicholas.lee, sidjha, sehoonkim, rtabrizi, suhong.moon, chooper, gopala, keutzer, amirgh\}@berkeley.edu}}}
}

\begin{document}
\maketitle
\input{_s0_abstract.tex}
\input{_s1_intro.tex}
\input{_s2_related_work.tex}
\input{_s3_methods.tex}

\input{_s5_discussions.tex}
\input{_s6_conclusion.tex}

\input{_s100_limitations.tex}

\section*{Acknowledgements}
We would like to thank Apple for sponsoring this project, as well as support from Microsoft through Accelerating Foundation Models Research Program. We also thank Sunjin Choi for his insights in energy cost associated with local and cloud deployment. Our conclusions do not necessarily reflect the position or the policy of our sponsors, and no official endorsement should be inferred.
\bibliography{tinyagent}
\clearpage

\appendix

\end{document}

%% file: packages.tex
\usepackage[preprint]{acl}

\usepackage{times}
\usepackage{latexsym}

\usepackage[T1]{fontenc}

\usepackage[utf8]{inputenc}

\usepackage{microtype}

\usepackage{inconsolata}

\usepackage{graphicx}

\usepackage[utf8]{inputenc} 
\usepackage[T1]{fontenc}    
\usepackage{hyperref}       
\usepackage{url}            
\usepackage{booktabs}       
\usepackage{amsfonts}       
\usepackage{nicefrac}       
\usepackage{microtype}      
\usepackage{xcolor}         
\usepackage{booktabs}
\usepackage{multicol}
\usepackage{enumitem}

\newcommand\hb{ \rowcolor{teal!13}}

\usepackage{subfig}
\usepackage{psfrag}
\usepackage{verbatim}
\usepackage{mathrsfs}
\usepackage{amssymb}%
\usepackage{pifont}%
\usepackage{multirow}
\usepackage{longtable}
\usepackage{listings}
\usepackage{graphicx}
\usepackage{amsmath}
\usepackage{mathtools}
\usepackage{amsthm}
\usepackage{xspace}
\usepackage[subtle]{savetrees}
\usepackage[most]{tcolorbox}
\usepackage{colortbl}

\newcommand{\bluecella}[1]{\cellcolor{teal!8}#1}
\newcommand{\bluecellb}[1]{\cellcolor{teal!20}#1}

\newcommand\blfootnote[1]{%
  \begingroup
  \renewcommand\thefootnote{}\footnote{#1}%
  \addtocounter{footnote}{-1}%
  \endgroup
}

%% file: _s0_abstract.tex
\begin{abstract}
Recent large language models (LLMs) have enabled the development of advanced agentic systems that can integrate various tools and APIs to fulfill user queries through function calling. However, the deployment of these LLMs on the edge has not been explored since they typically require cloud-based infrastructure due to their substantial model size and computational demands.
To this end, we present TinyAgent, an end-to-end framework for training and deploying task-specific small language model agents capable of function calling for driving agentic systems at the edge. We first show how to enable accurate function calling for open-source models via the LLMCompiler framework. 
We then systematically curate a high-quality dataset for function calling, which we use to fine-tune two small language models, TinyAgent-1.1B and 7B.  
For efficient inference, we introduce a novel tool retrieval method to reduce the input prompt length and utilize quantization to further accelerate the inference speed. 
As a driving application, we demonstrate a local Siri-like system for Apple's MacBook that can execute user commands through text or voice input. \blfootnote{*Equal contribution}
Our results show that our models can achieve, and even surpass, the function-calling capabilities of larger models like GPT-4-Turbo, while being fully deployed at the edge. We open-source our dataset, models, and installable package\footnote{\url{https://github.com/SqueezeAILab/TinyAgent}} and provide a demo video for our MacBook assistant agent\footnote{\url{https://www.youtube.com/watch?v=0GvaGL9IDpQ}}.
\end{abstract}

%% file: _s1_intro.tex
\section{Introduction}


The ability of LLMs to execute commands through plain language (e.g. English) has enabled agentic systems that can complete a user query by orchestrating the right set of tools (e.g. ToolFormer \cite{schick2024toolformer}, Gorilla \cite{patil2023gorilla}). This, along with the recent multi-modal efforts such as the GPT-4o~\cite{openai_gpt4o} or Gemini-1.5~\cite{google_gemini_2024}, has expanded the realm of possibilities with AI agents. 
However, the large model size and computational requirements of these models often requires their inference to be performed on the cloud. This can create several challenges for their widespread adoption.
First, uploading data such as video, audio, or text documents to a third-party vendor on the cloud, can result in privacy issues. Second, this requires cloud/Wi-Fi connectivity which is not always possible. For instance, a robot deployed in the real world may not always have a stable connection. Besides that, latency could also be an issue as uploading large amounts of data to the cloud and waiting for the response could slow down response time, resulting in unacceptable time-to-solution. These challenges could be solved if we deploy the LLM models locally at the edge.

Current LLMs like GPT-4o \cite{openai_gpt4o} or Gemini-1.5 \cite{google_gemini_2024} are too large for local deployment. One contributing factor is that a lot of the model size ends up memorizing general information about the world into its parametric memory which may not be necessary for a specialized downstream application.
For instance, if you ask a general factual question to these models like a historical event or well-known figures, they can produce the results using their parametric memory, even without having additional context in their prompt. 
This implicit memorization of training data into the parametric memory might be correlated with “emergent” phenomena in LLMs such as in-context learning and complex reasoning, which has been the driving force behind scaling the model size. 

This leads to an intriguing research question:

\textit{Can a smaller language model with significantly less parametric memory emulate such emergent ability of these larger language models?}

In this work, we demonstrate that this is feasible by training smaller models with specialized, high-quality data that does not require recalling generic world knowledge. 
Our goal is to develop Small Language Models (SLMs) that can be securely and privately deployed at the edge while maintaining the complex reasoning capability to understand natural language queries and orchestrate tools and APIs to accomplish user commands.

To achieve this, we first explore enabling small open-source models to perform accurate function calling, a key component of agentic systems. 
Off-the-shelf SLMs often lack sophisticated function calling capabilities and require fine-tuning.
Next, we discuss systematically curating high-quality function calling datasets to train these SLMs, using a specialized Mac assistant agent as our primary application. 
We demonstrate that fine-tuning the models on this curated dataset can enable SLMs to exceed GPT-4-Turbo’s function calling performance. 
Finally, we enhance the inference efficiency of these fine-tuned models using a novel Tool RAG method and quantization, allowing for efficient edge deployment with real-time responses.

%% file: _s2_related_work.tex
\section{Related Work}
\subsection{Function Calling LLMs}
The sophisticated reasoning capabilities of recent LLMs have enabled them to call functions (i.e., tools), where LLMs determine which function to invoke among user-provided functions along with the associated arguments. 
This allows LLMs to use external functions (e.g. calculators or search engines) to provide more accurate answers to user queries than by responding directly.
A pioneering work in this area is Toolformer~\cite{schick2024toolformer}, which has inspired various tool-calling frameworks~\cite{ruan2023tptu,shen2024hugginggpt,liang2024taskmatrix}.
ReAct~\cite{yao2022react} introduced a reasoning-and-action process that improved LLMs' interaction with external environments, which has become a back-bone for different open-source frameworks~\cite{Liu_LlamaIndex_2022, langchain}.
More recently, Gorilla~\cite{patil2023gorilla} and ToolLLM~\cite{qin2023toolllm} have demonstrated that an open-source LLM can be fine-tuned to obtain function-calling capabilities in diverse real-world use cases.
One noticeable work is Octopus~\cite{chen2024octopus} which introduces on-device LLMs that invoke software APIs.
TinyAgent pushes this boundary by enabling efficient inference via parallel function calling~\cite{kim2023llm} as well as a novel tool retrieval method, similar to~\cite{moon2024efficient}.
Furthermore, our method does not require any architectural changes, making it compatible with a wider range of open-source models.

\subsection{Dataset Synthesis}
To address the problem of not having enough data for finetuning, a popular method has emerged to use LLMs to synthesize new training datapoints \cite{deng2023rephrase, prasad2023rephrase, fu2023specializing, dai2023auggpt, ubani2023zeroshotdataaug, 
fang2023using, liu2023tinygsm, yu2023metamath, kumar2020data, yoo2021gpt3mix,wang2022self,lee2024llm2llm}.
While these techniques create very good results, they often generate a significant amount of training data. Recent advancements have shown that by filtering these datasets or generating smaller, higher quality datasets, one can achieve similar or better performance ~\cite{chen2023alpagasus, cao2023instruction, wei2023instructiongpt4, zhou2023lima}. TinyAgent builds on these works by constructing a pipeline that systematically generates high-quality, task-specific function-calling datasets, ensuring efficient training and robust performance even with smaller, curated datasets.

\subsection{Device Control}
Recent advancements in device control have introduced large-scale benchmarks and datasets focused on the Android environment \cite{rawles2024androidinthewild, zhang2024androidzoochainofactionthoughtgui, rawles2024androidworlddynamicbenchmarkingenvironment, lee2024benchmarking}, which explore UI-based agents with low-level controls such as typing, scrolling, and tapping. They are primarily concerned with mobile device interactions in simulated environments, but they do not address the challenges of deploying small language models directly on the device, which is crucial for real-world applications where cloud resources are unavailable or impractical. More recently, UFO \cite{zhang2024ufouifocusedagentwindows} introduced a dual-agent framework that leverages vision and language to enable UI-focused agents to operate within Windows OS applications. However, similar to earlier works, UFO also focuses on low-level control mechanisms and does not address the deployment of small language models directly on the device. TinyAgent pushes this boundary by formulating device control as a high-level function-calling problem instead of low-level UI actions, utilizing task-specific abstractions that allow for more robust and efficient execution of commands. By running fully locally on MacOS, TinyAgent offers a more realistic and practical solution for device control, making it well-suited for real-life scenarios where on-device deployment is necessary.

%% file: _s3_methods.tex
\section{TinyAgent}

\begin{figure*}[!t]
    \centering
    \captionsetup{}
    \includegraphics[width=0.6\linewidth]{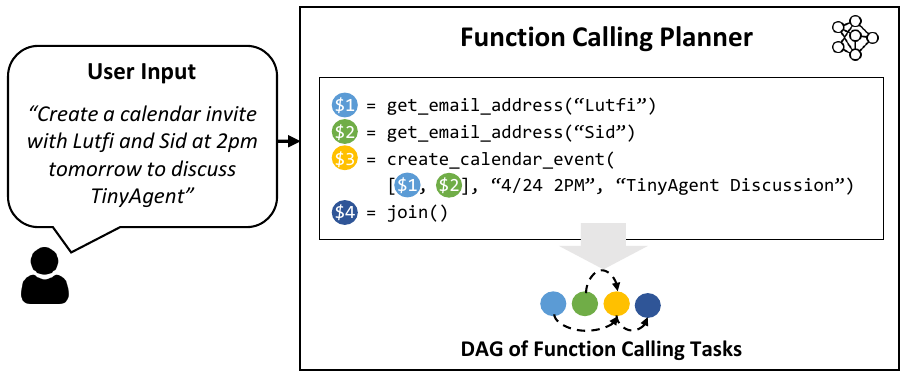}
    \caption{Overview of the LLMCompiler Function Calling Planner. The Planner understands the user query and generates a sequence of tasks with their inter-dependencies. These tasks are then dispatched by the LLMCompiler framework to accomplish the user command. In this example, Task \$1 and \$2 are fetched together to retrieve the email addresses of Sid and Lutfi independently. After each task is performed, the results are forwarded to Task \$3 which creates the calendar event. Before executing Task \$3, LLMCompiler replaces the placeholder variables (e.g., the variable \$1 and \$2 in Task \$3) with actual values.} 
    \label{fig:function_calling}
\end{figure*}

\subsection{Teaching LLMs to do Function Calling}

As mentioned above, our main interest is applications where the AI agent translates the user query into a sequence of function calls to complete the tasks. In such applications, the model does not need to write the function definition itself since the functions (or APIs) are mostly pre-defined and already available. Therefore, what the model needs to do is to determine (i) which functions to call, (ii) the corresponding input arguments, and (iii) the right order of calling these functions (i.e. function orchestration) based on the required interdependency across the function calls.

The first question is to find an effective way to equip SLMs to perform function calling. Large models such as GPT-4 are able to perform function calling, but how can this be achieved with open source models? LLMCompiler \cite{kim2023llm} is a recent framework that enables this by instructing the LLM to output a function calling plan that includes the set of functions that it needs to call along with the input arguments and their dependencies (see the example in Figure~\ref{fig:function_calling}). Once this function calling plan is generated, we can parse it and call each function based on the dependencies.

The critical part here is how to teach the model to create this function calling plan with the right syntax and dependency
. The original LLMCompiler~\cite{kim2023llm} only considered large models, such as LLaMA-2 70B \cite{touvron2023llama}, which have complex reasoning capabilities to create the plan when provided with sufficient instructions in their prompts. 
Unfortunately, our initial experiments showed that off-the-shelf small models such as TinyLlama-1.1B \cite{zhang2024tinyllama} (or even the larger Wizard-2-7B model \cite{huggingface_notwizardlm}) are not able to output the correct plans when prompted the same way.
The errors ranged from problems such as using the wrong set of functions, hallucinated names, wrong dependencies, and inconsistent syntax.

\begin{figure*}[!t]
    \centering
    \captionsetup{}
    \includegraphics[width=0.9\linewidth]{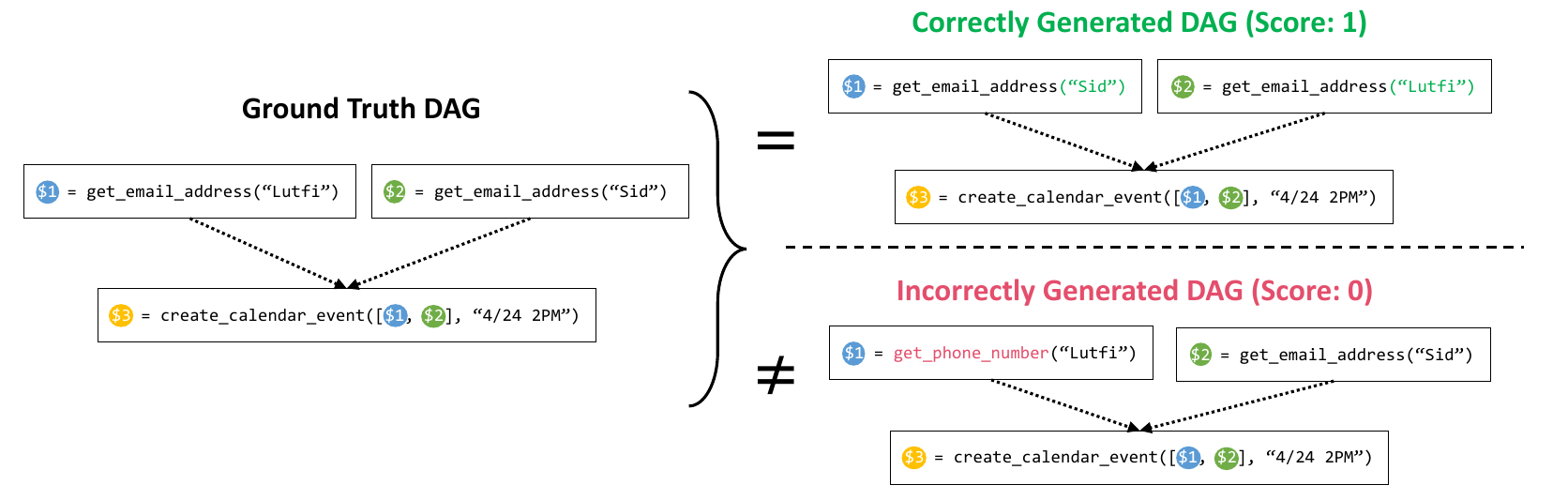}
    \vspace{-1mm}
    \caption{Graph Isomorphism Success Rate. The model scores a success rate of 1 only if the DAG of its generated plan is isomorphic to the DAG of the ground truth plan; and 0 otherwise. In the above example, for the top case, although the order of the \texttt{get\_email\_address} calls are different from the ground truth plan (the ground truth plan gets the email address of Lutfi before Sid, and the generated plan gets the email address of Sid before Lutfi), since the two DAGs are isomorphic to each other, the plan gets 1 success rate. For the bottom case, since the predicted DAG contains a wrong node, corresponding to a wrong function call, the plan gets 0 success rate.} 
    \label{fig:dag_score}
    \vspace{2mm}
\end{figure*}

This is rather expected because these small models have been trained on generic datasets and primarily targeted to achieve good accuracy on general benchmarks which mostly test the model’s world knowledge and general reasoning or basic instruction following capability. To address this, we explored if fine-tuning these models on a high-quality dataset specially curated for function calling and planning can improve the accuracy of these small language models for a targeted task, potentially outperforming larger models. 
In Section~\ref{subsec:dataset}, we first discuss how we generated such a dataset, and then we discuss the fine-tuning approach in Section~\ref{subsec:finetuning}.

\subsection{Dataset Generation}
\label{subsec:dataset}
As a driving application, we consider a local agentic system for Apple’s Macbook that solves user’s day-to-day tasks.
Particularly, the agent is equipped with 16 different functions that can interact with different applications on Mac, which includes:
    \vspace{-2.5mm}

\begin{itemize}[leftmargin=3mm]
    \item \textbf{Email:} Compose a new email or reply to/forward emails
    \vspace{-2.5mm}
    \item \textbf{Contacts:} Retrieve phone numbers or email addresses from the contacts database
        \vspace{-2.5mm}
    \item \textbf{SMS:} Send text messages to contact(s)
        \vspace{-2.5mm}
    \item \textbf{Calendar:} Create calendar events with details such as title, time, attendees, etc.
        \vspace{-2.5mm}
    \item \textbf{Notes:} Create, open, or append content to notes in various folders
        \vspace{-2.5mm}
    \item \textbf{Reminder:} Set reminders for various activities and tasks
        \vspace{-2.5mm}
    \item \textbf{File management:} Open, read, or summarize documents in various file paths
        \vspace{-2.5mm}
    \item \textbf{Zoom meetings:} Schedule and organize Zoom meetings
\end{itemize}

Predefined Apple scripts exist for each of these functions/tools, and all that the model needs to do is to take advantage of the predefined APIs and determine the right function calling plan to accomplish a given task, such as in Figure~\ref{fig:function_calling}. 
However, as discussed previously, we need a dataset for training and evaluating SLMs since their off-the-shelf function calling capability is subpar.

Creating handcrafted data with diverse function calling plans is both challenging and not scalable. However, we can curate synthetic data using a powerful LLM like GPT-4-Turbo. 
Such an approach is becoming a common method where a capable LLM is instructed to generate data similar to a given set of sample examples or templates.
In our work, we used a similar approach, but instead of providing the LLM with generic user queries as templates, we provide it with various sets of functions and instruct it to generate realistic user queries that require those functions to accomplish the task, along with the associated function calling plan and input arguments, like the example shown in Figure~\ref{fig:function_calling}. To verify the validity of the generated data, we incorporated sanity checks on the function calling plan to make sure that they form a feasible graph, and that the function names and input argument types are correct. With this approach, we created 80K training data, 1K validation data, and 1K testing data, with a total cost of only $\sim$\$500.

\begin{figure*}[!t]
    \centering
    \captionsetup{}
    \includegraphics[width=0.75\linewidth]{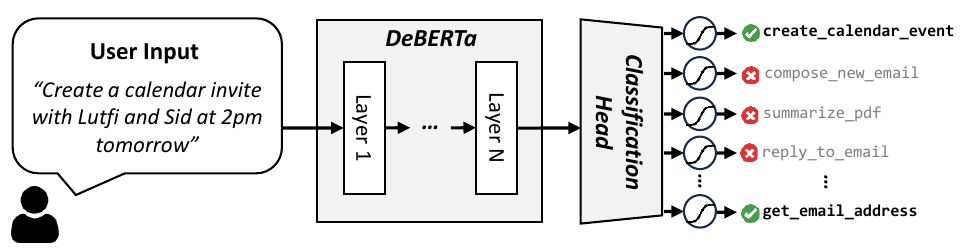}
    \caption{Overview of our Tool RAG scheme. We formulate tool retrieval as a multi-label classification problem. The user query is given as input to the fine-tuned DeBERTa-v3-small model, which outputs a 16-dimensional vector indicating tool probabilities. Tools with probabilities higher than 50\% are selected, averaging 3.97 tools per query compared to 6 tools in basic RAG.} 
    \label{fig:tool_rag}
\end{figure*}
\begin{figure}[!t]
    \centering
    \captionsetup{}
    \includegraphics[width=\linewidth]{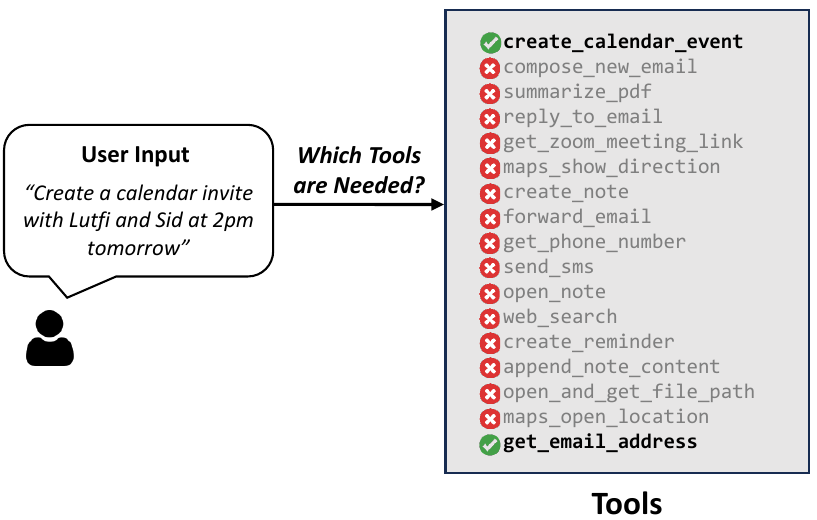}
    \caption{ Efficient tool selection based on a user input. Not all user inputs require all available tools; hence, it is imperative to select the right set of tools to minimize the prompt size and increase performance. In this case, the LLM only needs the functions that get email addresses and create a calendar event  to accomplish its task.} 
    \label{fig:tools}
\end{figure}

\subsection{Fine-tuning for Improved Function Calling Reasoning}
\label{subsec:finetuning}
With our dataset in place, we can now proceed to fine-tune off-the-shelf SLMs to enhance their function calling capability. We started with two base small models: TinyLlama-1.1B (instruct-32k) and Wizard-2-7B. 
For fine-tuning these models, we first need to define a metric to evaluate their performance. 
Our objective is for these models to accurately generate the right plan, i.e., to select the right set of functions \textit{and} to orchestrate them in the right order. Therefore, we define a success rate metric that assigns 1 if both criteria are met, and 0 otherwise. Checking whether the model has selected the right set function calls is straightforward. To additionally ensure that the orchestration of these functions is correct, we construct a Directed Acyclic Graph (DAG) of the function calls based on the dependencies, as shown in Figure~\ref{fig:dag_score}, where each node represents a function call and a directed edge from node A to B represents their interdependency (i.e. function B can only be executed after the execution of function A). Then we compare if this DAG is identical to that of the ground truth plan to verify the accuracy of the dependencies.

After defining our evaluation metric, we applied LoRA \cite{hu2021lora} to fine-tune the models for 3 epochs using a learning rate of 7e-5 over the 80K training examples, and selected the best checkpoint based on validation performance. For fine-tuning, our prompt included not only the descriptions of the ground truth functions (i.e. functions used in the ground truth plan) but also other irrelevant functions as negative samples. We found the negative samples to be particularly effective for teaching the model how to select appropriate tools for a given query, hence improving the post-training performance. Furthermore, we also include several in-context examples demonstrating how queries are translated into a function calling plans. These in-context examples are selected through a Retrieval Augmented Generation (RAG) process based on the user query from the data in the training dataset.

Using the above settings, we fine-tuned TinyLlama-1.1B/Wizard-2-7B models. After fine-tuning, the 1.1B model improved the success rate from 12.71\% to 78.89\%, and the 7B model performance improved from 41.25\% to 83.09\%, which is $\sim$4\% higher than GPT-4-Turbo.

\subsection{Efficient Inference with Tool RAG}

Our primary goal is to be able to deploy the TinyAgent model locally on a Macbook, which has limited computational and memory resources available as compared to the GPUs that closed-source models like GPT are deployed on. To achieve efficient performance with low latency we need to ensure that not only is the model size small, but that the input prompt is as concise as possible. The latter is an important contributor to latency and computational resource consumption due to the quadratic complexity of attention on sequence length.

The fine-tuned TinyAgent model discussed previously was fine-tuned with the description of all available tools in its prompt. However, we can significantly reduce the prompt size by only including the description of relevant tools based on the user query. For instance, consider the example shown in Figure~\ref{fig:tools} above, where the user is asking to create a calendar invite with two people. In this case, the LLM only needs the functions that get email addresses and create a calendar event in its prompt.

\begin{table*}[!t]
\vspace{-3mm}
\caption{
Comparison of TinyAgent performance with DeBERTa to Basic RAG and no RAG settings.
For Basic RAG, we retrieved top-3 most relevant tools. For our fine-tuned DeBERTa-v3-small model, we retrieved tools with a probability greater than 50\%, which retrieves $\sim$3.97 tools per query.}
\vspace{-1mm}
\begin{center}
\small{
\setlength{\tabcolsep}{6pt}{
\begin{tabular}{c|c|c|c|c}
\toprule
 \multirow{2}{*}{Tool RAG Method} & \multirow{2}{*}{Tool Recall} & Prompt Size & TinyAgent 1.1B 	& TinyAgent 7B \\
 & & (Tokens) & Success Rate (\%) & Success Rate (\%)  \\
\midrule
No RAG (all tools in the prompt) & 1 & 2762 & 78.89 & 83.09 \\
Basic RAG & 0.949 & 1674 & 74.88 & 78.50 \\
\hb Fine-tuned DeBERTa-v3-small (Ours) & \textbf{0.998} & \textbf{1397} & \textbf{80.06} & \textbf{84.95} \\

\bottomrule
\end{tabular}
}
}
\end{center}
\label{table:t1}
\vspace{2mm}
\end{table*}

\begin{table*}[!t]
\vspace{-3mm}
\caption{
Latency, size, and success rate of TinyAgent models before and after quantization. Latency is the end-to-end latency of the function calling planner, including the prompt processing time and generation.}
\vspace{-1mm}
\begin{center}
\small{
\setlength{\tabcolsep}{6pt}{
\begin{tabular}{c|c|c|c|c}
\toprule
Model &	Weight Precision &	Latency (seconds)	& Model Size (GB)	& Success Rate (\%) \\
\midrule
GPT-3.5 & Unknown & 3.2 & Unknown & 65.04 \\
GPT-4-Turbo & Unknown & 3.9 & Unknown & 79.08 \\
\midrule
 \multirow{2}{*}{TinyAgent-1.1B} & \bluecella{16} & \bluecella{3.9} & \bluecella{2.2} & \bluecella{80.06} \\
 & \bluecellb{4} & \bluecellb{\textbf{2.9}} & \bluecellb{\textbf{0.68}} & \bluecellb{\textbf{80.35}} \\
 \midrule
 \multirow{2}{*}{TinyAgent-7B} & \bluecella{16} & \bluecella{19.5} & \bluecella{14.5} & \bluecella{84.95} \\
& \bluecellb{4} & \bluecellb{\textbf{13.1}} & \bluecellb{\textbf{4.37}} & \bluecellb{\textbf{85.14}} \\

\bottomrule
\end{tabular}
}
}
\end{center}
\label{table:t2}
\vspace{2mm}
\end{table*}

To take advantage of this observation, we need to determine which functions are required to accomplish the user’s command, which we refer to as Tool RAG given its similarity with how RAG works. However, the model performs poorly when we use a basic RAG method where we retrieve the relevant tools based on the embedding similarity of the user query and the tools. 
This is because completing a user’s query often requires using several auxiliary tools which may be missed with a simple RAG method if the embedding of the auxiliary tool is not similar to the user query. For instance, the example shown in Figure~\ref{fig:tools} requires calling get\_email\_address function even though the user query is just asking about creating a calendar invitation.

This can be addressed by treating the problem as a classification of which tools are needed. To that end, we fine-tuned a DeBERTa-v3-small \cite{he2021debertav3} model on the training data to perform a 16-way classification as shown in Figure~\ref{fig:tool_rag}. The user query is given as an input to this model, and then we pass the CLS token at the end through a simple fully connected layer of size 768x16 to transform it into a 16 dimensional vector (which is the total size of our tools). The output of this layer is passed through a sigmoid layer to produce the probability of selecting each tool. During inference, we select the tools that have probably higher than 50\%, and if so, we include their description in the prompt. On average we noticed that only 3.97 tools are retrieved with a recall of 0.998, whereas the basic RAG requires using the top 6 tools to achieve a tool recall of 0.968.

We evaluated the model performance after incorporating Tool RAG. The results are shown in Table~\ref{table:t1}, where we report the performance of the simple RAG system along with the fine-tuned DeBERTa approach. As one can see, the DeBERTa based Tool RAG method achieves almost perfect recall performance, improves the baseline accuracy, while reducing the prompt size by $\sim$2x tokens.

\subsection{Fast Edge Deployment with Quantization}

Deploying models at the edge, such as on consumer MacBooks, can still be challenging even for small models with O(1B) parameters, since loading the model parameters can consume a large portion of the available memory. A solution to these issues is quantization, which allows us to store the model at a reduced bit precision. Quantization not only reduces the storage requirements and model footprint, but also cuts down the time and resources needed to load model weights into memory, thereby reducing the overall inference latency as well.
For more information on quantization, refer to~\cite{gholami2022survey}.

To more efficiently deploy the models, we quantized the models into 4-bit with a group size of 32, which is supported by the llama.cpp framework with quantization-aware training. As shown in Table~\ref{table:t2}, the 4-bit models result in 30\% better latency, along with a 4x reduction in the model size. We also notice slight accuracy improvement which is due to the additional fine-tuning with simulated quantization.

%% file: _s5_discussions.tex
\section{Putting It All Together}
We provide a demo video of the final TinyAgent-1.1B model deployed on a Macbook Pro M3\footnote{\url{https://www.youtube.com/watch?v=0GvaGL9IDpQ}}, which can be downloaded and tested on Mac from the link\footnote{\url{https://github.com/SqueezeAILab/TinyAgent/raw/main/TinyAgent.zip}}.
It not only runs all of the model inference locally on your computer, but it also allows you to provide commands through audio. We process the audio locally as well using the Whisper-v3~\cite{radford2022robust} model from OpenAI deployed locally using the whisper.cpp framework. The greatest surprise for us was that the accuracy of the 1.1B model exceeds that of GPT-4-Turbo, and is markedly fast while deployed locally and privately on-device.

%% file: _s6_conclusion.tex
\section{Conclusions}

To summarize, we introduced TinyAgent and showed that it is indeed possible to train a small language model and use it to power a semantic system that processes user queries. In particular, we considered a Siri-like assistant for Mac as a driving application. The key components for enabling it is to (i) teach off-the-shelf SLMs to perform function calling through LLMCompiler framework, (ii) curate high quality function calling data for the task at hand, (iii) fine-tune the off-the-shelf model on the generated data, and (iv) enable efficient deployment by optimizing the prompt size through only retrieving the necessary tools based on the user query through Tool RAG, as well as quantized model deployment to reduce inference resource consumption. After these steps, our final models achieved 80.06\% and 84.95\% for the TinyAgent-1.1.B and 7B models which exceed GPT-4-Turbo’s success rate of 79.08\% on this task.

%% file: _s100_limitations.tex
\section{Ethics Statement}

Deploying TinyAgent to operate agentic systems at the edge presents several ethical considerations that are integral to our design and operational philosophy.

\textbf{Accessibility and Inclusivity:} Ensuring that TinyAgent serves all users equitably, including those with disabilities, is a priority. We are committed to designing interfaces that are universally accessible, incorporating features such as voice recognition that can understand diverse speech patterns and text-to-speech technologies that are clear and easily comprehensible. Further, we are exploring adaptive technologies that can adjust to the specific needs of users with varying abilities, ensuring that everyone can benefit from TinyAgent’s capabilities without barriers.

\textbf{Human Oversight:} While TinyAgent demonstrates robust capabilities in function calling, the risk of hallucination and erroneous responses by LLMs remains \cite{zhang2023siren}. To mitigate this, it is essential to maintain human oversight throughout the operational loop, not just at the endpoint. This means integrating mechanisms for regular checks and balances where humans can review, override, or refine decisions made by TinyAgent. Future iterations of our system will aim to facilitate even more seamless human-agent collaboration to enhance decision accuracy and reliability.

\textbf{Cultural and Bias Considerations:} 
Synthetic datasets generated using simple or naive prompts often carry inherent biases, such as those related to regional or cultural specificity \cite{yu2024large}. Because task-specific agent systems like TinyAgent rely on synthetic data, their effectiveness and impartiality can be impacted when operating across different demographic landscapes. In response, we integrate diverse cultural data and demographic groups in our data generation processes to mitigate these biases. Our aim is to ensure that the synthetic data fueling TinyAgent is as inclusive and unbiased as possible, supporting a function-calling system that is culturally aware and equitably serves a global user base.